\documentclass[10pt,twocolumn,letterpaper]{article}

\usepackage[pagenumbers]{cvpr} % To force page numbers, e.g. for an arXiv version

\definecolor{cvprblue}{rgb}{0.21,0.49,0.74}
\usepackage[most]{tcolorbox} 
\usepackage{newtxmath}
\usepackage{tablefootnote}
\usepackage{multirow}
\usepackage{textcomp} % for \textdagger

\definecolor{Embedlens}{HTML}{66bdbb}
\definecolor{Logitlens}{HTML}{fd757a}
\definecolor{deadTokens}{HTML}{dd4b4e}
\definecolor{aliveTokens}{HTML}{66bdbb}
\definecolor{sinkTokens}{HTML}{515a85}

\usepackage[pagebackref,breaklinks,colorlinks,allcolors=cvprblue]{hyperref}
\usepackage{pifont}
\usepackage{booktabs} % for professional tables
\usepackage[table]{xcolor}

\def\paperID{*****} % *** Enter the Paper ID here
\def\confName{CVPR}
\def\confYear{2026}

\title{What Do Visual Tokens Really Encode? Uncovering Sparsity and Redundancy in Multimodal Large Language Models}

\author{
Yingqi Fan\textsuperscript{1,2}\hspace{0.2cm}
Junlong Tong\textsuperscript{1,2,3}\hspace{0.2cm}
Anhao Zhao\textsuperscript{1,4}\hspace{0.2cm}
Xiaoyu Shen\textsuperscript{1,2}\thanks{Corresponding author.} \\
\textsuperscript{1}Institute of Digital Twin, Eastern Institute of Technology, Ningbo \\
\textsuperscript{2}Ningbo Key Laboratory of Spatial Intelligence and Digital Derivative\\
\textsuperscript{3}Shanghai Jiao Tong University \quad
\textsuperscript{4}The Hong Kong Polytechnic University \\ 
\texttt{\small yingqi949@gmail.com  xyshen@eitech.edu.cn}
}

\begin{document}
\maketitle
\begin{abstract}
Multimodal large language models (MLLMs) project visual tokens into the embedding space of language models, yet the internal structuring and processing of visual semantics remain poorly understood. In this work, we introduce a two-fold analytical framework featuring a novel probing tool, \textbf{EmbedLens}, to conduct a fine-grained analysis. We uncover a pronounced semantic sparsity at the input level: visual tokens consistently partition into sink, dead, and alive categories. Remarkably, only the alive tokens, comprising $\approx60\%$ of the total input, carry image-specific meaning. Furthermore, using a targeted patch-compression benchmark, we demonstrate that these alive tokens already encode rich, fine-grained cues (e.g., objects, colors, and OCR) prior to entering the LLM. Internal visual computations (such as visual attention and feed-forward networks) are redundant for most standard tasks. For the small subset of highly vision-centric tasks that actually benefit from internal processing, we reveal that alive tokens naturally align with intermediate LLM layers rather than the initial embedding space, indicating that shallow-layer processing is unnecessary and that direct mid-layer injection is both sufficient. Ultimately, our findings provide a unified mechanistic view of visual token processing, paving the way for more efficient and interpretable MLLM architectures through selective token pruning, minimized visual computation, and mid-layer injection.~\footnote{The code is released at: \href{https://github.com/EIT-NLP/EmbedLens}{https://github.com/EIT-NLP/EmbedLens}.}

\end{abstract}    
\section{Introduction}
\label{sec:intro}

Multimodal large language models (MLLMs) increasingly underpin visual–language applications by projecting image features into the embedding space of powerful language models~\citep{clip, llava, Qwen2.5-VL}. In practice, a vision encoder, often pre-trained with contrastive learning (e.g., CLIP~\citep{clip}), produces a sequence of patch embeddings, and a projector maps these features into the LLM's embedding space~\cite{blip2,lin2024preserve}. This design has enabled impressive progress, but it also exposes a structural tension: contrastive pretraining encourages global image–text alignment, while the LLM ingests information as a sequence of local, patch-level tokens~\cite{yin2024survey}.

This discrepancy creates a crucial but largely unexplored gap in our understanding. How is the globally-aligned semantic information distributed across the local tokens? Do all patches carry meaningful semantics? Do patch embeddings already encode discrete, language-like concepts (``pre-linguistic'') that LLMs can directly digest, or do they require extensive LLM processing to become understandable? Answering these questions is essential for building more efficient, interpretable, and powerful MLLMs.

To investigate this, we begin by analyzing the visual token embeddings at the LLM's input layer. Our methodology is two-fold: we first apply similarity-based clustering to map the tokens' macro-level structural organization and then introduce \emph{EmbedLens}, a fine-grained semantic probing framework we developed to dissect the micro-level semantic attributes encoded within each token and cluster.

Our empirical analyses  reveal a striking discovery: the visual tokens entering the LLM are far from homogeneous. Instead, they partition into three functionally distinct groups with stable, cross-image properties: (1) \textbf{Sink Tokens:} Image-agnostic tokens whose embeddings remain nearly identical regardless of the input. They serve a purely \emph{structural role} (\eg, stabilizing attention distributions) but carry no image-specific semantics.
(2) \textbf{Dead Tokens:} Similarly image-independent, but unlike sink tokens, they attract minimal attention and exert negligible influence on model computation, serving \emph{neither a structural nor a contextual purpose}.
(3) \textbf{Alive Tokens:} Tokens that cluster near the text semantic center. These serve a clear \emph{contextual role}, acting as the primary interface for translating visual content into language-compatible semantics.

Under this partitioning, we reveal \emph{a pronounced semantic sparsity: a large fraction ($\approx$40\%) of all visual tokens are sink or dead tokens}. We demonstrate that they can be safely removed without performance loss, and in some cases, even improve performance by removing distractions.

Having established this sparsity, we narrow our focus to the remaining 60\% alive tokens, which serve as the sole carriers of image-specific semantics. This raises two consecutive questions: (1) How much semantic information do these alive tokens encode \emph{before} entering the LLM? and (2) Once injected, what role does the LLM's internal visual computation (\ie, visual self-attention and feed-forward networks (FFN)) play in further processing them?

To assess their initial information capacity, we construct a targeted patch-compression benchmark that forces object or OCR information into a single visual patch. We find that a single alive token routinely bundles multiple semantic attributes (such as object identity, color, shape, and OCR glyphs). They function as highly dense information units that are largely ``pre-linguistic,'' requiring little to no further translation to be understood by the text model.

Given this strong pre-linguistic alignment, we next examine the necessity of the LLM's internal visual processing. Surprisingly, for most standard tasks (e.g., general VQA, OCR, and hallucination mitigation), entirely bypassing the visual-only FFN and self-attention layers has a negligible impact on performance, and can occasionally even improve it. We observe that further internal processing can actually introduce biases, such as rendering color predictions overly dependent on background context. This demonstrates that \emph{the projector already aligns alive tokens so effectively that they can be ``read'' by the LLM's cross-attention mechanisms without needing further visual transformation}. Ultimately, only a small subset of vision-centric tasks truly benefits from this internal processing.

Finally, since only a subset of vision-centric tasks requires meaningful LLM-internal processing, we investigate \emph{where} in the network depth this useful processing occurs. We find that the vector norm of alive tokens naturally align with representations in the middle layers of the LLM rather than the initial embedding space. The shallowest layers perform little transformation and can even degrade performance if visual tokens are forced through them. This indicates that projectors intentionally map visual embeddings closer to mid-layer representations, implying that \emph{directly injecting visual tokens into middle layers is sufficient}.

Together, these findings provide a unified understanding of how visual semantics are organized, propagated, and transformed inside MLLMs. Our main contributions are: (1) We propose \emph{EmbedLens}, a practical, model-agnostic probing framework for patch-level inspection of visual tokens, validated across multiple MLLM families (\eg LLaVA, Qwen-VL, InternVL variants). (2) We provide the first empirical evidence that visual tokens partition into sink, dead, and alive groups, revealing massive semantic sparsity at the input. (3) We demonstrate a two-fold redundancy in MLLMs: internal visual computation (FFN/self-attention) is unnecessary for most tasks, and shallow-layer processing is often detrimental, as tokens are aligned with middle layers. (4) We provide mechanistic insights that motivate token-efficient architectures, selective visual processing, and improved interpretability for multimodal systems.

\section{Motivation}
\label{sec:2_MOT}

% The contrastive pre-training objective of visual encoders (like CLIP) is to align global, image-level representations. However, downstream multimodal LLMs heavily rely on fine-grained, patch-level visual features to understand visual information.
% This fundamental granularity mismatch between the pre-training objective (global alignment) and the downstream application (local reasoning) raises an underexplored question: how are the semantics learned from global image-text alignment transformed, organized, and ultimately interpreted within the LLMs?

To bridge the fundamental mismatch between the\textit{ global alignment} of visual encoders and the \textit{local, patch-level} processing of MLLMs, we first investigate how global semantics are transformed and organized at the individual token level.
Given that the dominant MLLM paradigm employs a projection layer to bridge the visual encoder to the LLM's word embedding space, this study targets the post-projection visual embeddings as an entry point.
A preliminary investigation into this embedding space reveals a key finding: the visual tokens are far from a flat, undifferentiated set of features. Instead, they exhibit a non-trivial cluster structure. More specifically, by analyzing token positions and similarities, we identify three stable and distinct clustering patterns:\footnote{All observations are statistically derived from 10K samples from the COCO dataset~\cite{coco}. The detail can be found in the Appendix.}
cross-image high-homogeneity cluster, text-distancing cluster, and text-proximity cluster.

\begin{figure}[t]
    \centering
    \includegraphics[width=1\linewidth]{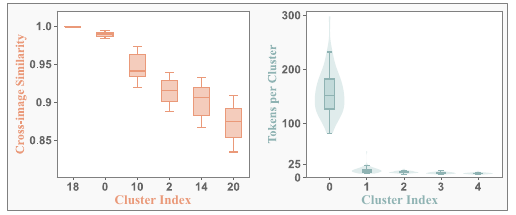}
    \vspace{-20pt}
    \caption{\small \textbf{Cluster dominance and Cross-image stability in post-projection visual embeddings}. (Left) Cluster similarity cross images. (Right) Number of tokens in the top-5 clusters. The cluster index is ranked by the number of tokens.}
    \label{fig:02_cluster}
\end{figure}
\begin{figure}[t]
    \centering
    \includegraphics[width=1\linewidth]{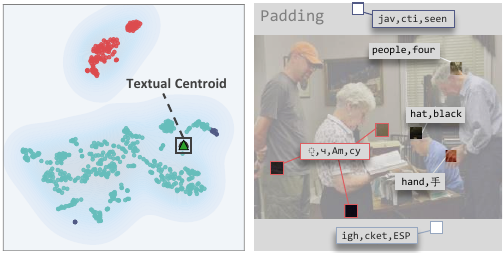}
    \vspace{-20pt}
    \caption{\small {(\textbf{Left}) The t-SNE visualization of visual and textual clusters in word embedding space. (\textbf{Right}) An example illustrating the multi-semantic properties of a token within the green-marked text-proximity cluster shown in the left figure.}}
    \label{fig:21_tSNE_semantics}
\end{figure}

\paragraph{Cross-image High-homogeneity Cluster.} We first identify a distinct set of tokens that not only cluster internally but also exhibit high homogeneity (i.e., high cosine similarity) across entirely different images, regardless of their content. 
As illustrated in left of Fig.~\ref{fig:02_cluster}, the cross-image similarity for this top-ranked cluster approaches 100\%. This cluster corresponds to the purple region in the left of Fig.~\ref{fig:21_tSNE_semantics}.
This potentially suggests they represent input-agnostic structural artifacts or representational priors of the projection layer, rather than image-specific semantics.

\paragraph{Text-distancing Cluster.} Moving to the intra-image structure, we observe a second distinct cluster of tokens that consistently maintains a high cosine distance from the text embeddings, potentially contributing minimally to semantic reasoning or performance. This cluster is visually identified as the red area in Fig.~\ref{fig:21_tSNE_semantics}. Furthermore, this cluster is exceptionally large and dominant, consistently containing far more tokens than all remaining clusters combined, as quantified by the cluster at index 0 in Fig.~\ref{fig:21_tSNE_semantics}.

\paragraph{Text-proximity Cluster.} In contrast to the distancing cluster, we find several smaller clusters grouped around the text centroid. These tokens exhibit a low cosine distance (i.e., high proximity) to the text representations, suggesting a greater potential for contributing to semantic understanding, as visually depicted in Fig.~\ref{fig:21_tSNE_semantics}. Building on this, we find that tokens corresponding to these clusters often exhibit multi-semantic properties, aligning with multiple distinct textual concepts (also illustrated in Fig.~\ref{fig:21_tSNE_semantics}).

The identification of these three stable and structurally distinct token archetypes reveals that visual tokens entering an MLLM cannot be treated as a uniform input stream. Instead, they exhibit markedly different positions, semantic potentials, and structural behaviors.
Building on this observation, our work goes beyond the naive discovery of structural patterns: we develop dedicated analytical tools and conduct a systematic investigation into how each token category participates in, and affects, the visual–language alignment process within MLLMs.

\section{\emph{EmbedLens}}
\label{sec:3_EmbedLens}

% To probe these questions, we require a method that can directly examine whether visual embeddings themselves encode interpretable semantics \emph{before} any linguistic transformation by the LLM. 
To probe the semantic function of the tokens within the three identified clusters, we require a method that can directly examine whether their visual embeddings intrinsically encode interpretable semantics as a direct result of the projection alignment.
While a large number of works apply the unembedding matrix to translate soft visual representations into textual tokens~\cite{wu2025semantichubhypothesislanguage,wang2025understandingknowledgeevolveslarge,neo2025interpretingvisualinformationprocessing,CircuitProbe,logitlens,jiang2025interpretingeditingvisionlanguagerepresentations}, this fails to identify whether those semantics are already intrinsic to the representations produced by the vision encoder and projection, or are instead injected by the language backbone. 

\subsection{Nearest Embedding Retrieval} 
\label{subsec:31_embedlens}
Building upon this motivation, we introduce \emph{EmbedLens}, which probes the semantic content of representations directly within the \emph{input embedding space}. The core idea is to measure the similarity between a given target embedding (\eg, a projected visual token or an intermediate hidden state) and all token embeddings in the model’s input vocabulary. Formally, let $\mathbf{W}_{E} \in \mathbb{R}^{\mathcal{T} \times d}$ denote the input embedding matrix, where each row $\mathbf{e}_i$ corresponds to the embedding of token $i$, and let $\mathbf{h} \in \mathbb{R}^{d}$ be the target representation (\eg a visual token after projection). EmbedLens computes the cosine similarity between $\mathbf{h}$ and every token embedding, and retrieves the top-$k$ tokens with the highest scores:

\begin{equation}
    \text{EmbedLens}(\mathbf{h}) = \operatorname*{TopK}_{i \in \mathcal{V}}
    \left(
    \frac{\mathbf{h}^\top \mathbf{e}_i }
    {\|\mathbf{h}\|_2 \, \|\mathbf{e}_i\|_2}
    \right).
    \label{eq:EmbedLens}
\end{equation}

\subsection{Interpretability Applications of EmbedLens}
\label{subsec:32_embedlens_application}
\paragraph{Semantic Identification.}
To assess whether \emph{EmbedLens} successfully captures relevant semantics, we perform a label–token matching experiment on the MS~COCO~2017 dataset~\cite{coco}.  
For each annotated object label $t^*$ in an image, we check whether it appears within the top-$5$ retrieved tokens $\{\hat{t}_1, \ldots, \hat{t}_5\}$ of any projected visual embedding.  
Formally, the matching accuracy at layer~$\ell$ is defined as:
\begin{equation}
A^{(\ell)} = \frac{1}{N} \sum_{j=1}^{N} 
\vmathbb{1}\!\left[t^*_j \in \text{EmbedLens}_{top_k}(\mathbf{h}^{(\ell)}_j)\right],
\end{equation}
where $\mathbf{h}^{(\ell)}_j$ denotes the $j$-th visual token representation.  

\begin{figure}[h]
    \centering
    \includegraphics[width=1\linewidth]{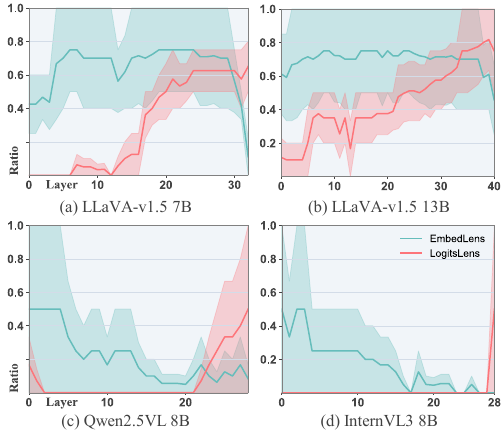}
    \vspace{-20pt}
    \caption{\small \textbf{Comparison between \textcolor{Embedlens}{EmbedLens} and \textcolor{Logitlens}{LogitLens} in label–token matching accuracy.} EmbedLens achieves higher semantic recall at shallow and middle layers across multiple MLLMs, suggesting that \emph{visual embeddings already encode object-level semantics prior to LLM fusion}.}
    \label{fig:03_NER_logitlens_compare}
\end{figure}

As shown in \cref{fig:03_NER_logitlens_compare}, \emph{EmbedLens} consistently identifies correct object labels in shallow and mid layers. Generalized across LLaVA-1.5-7B/13B, InternVL-3-8B, and Qwen2.5-VL-8B~\cite{Qwen2.5-VL,improvedllava,zhu2025internvl3}, this indicates that a substantial portion of \emph{object-level semantics can be recovered directly from visual embeddings, before any linguistic transformation}. 

\paragraph{Cluster Tracking.}
Since visual tokens are grouped by embedding similarity, tokens within the same cluster are expected to share consistent semantic characteristics. We apply \emph{EmbedLens} to each cluster’s centroid at model input to perform a discrete identification of its representative semantics by assigning a reference textual token $\hat t_
\mathbf{c}\in \textit{EmbedLens}_{k}(\mathbf{C})$. This provides an interpretable textual label for each visual cluster, enabling us to track specific clusters across different images and supporting cluster-level interpretability analysis.

With \emph{EmbedLens} providing semantic interpretation and cluster-level tracking, we next examine the high-homogeneity clusters to assess whether their repetitive or input-independent patterns correspond to redundant or sink-like representations.

% Having established a means to interpret the semantics of individual visual embeddings and locate the cluster, we next investigate their structural organization. Specifically, we perform an anchor-based clustering analysis to determine whether certain token groups encode repetitive or input-independent patterns—potentially corresponding to redundant or sink-like representations.

% Having established a means to interpret the semantics of individual visual embeddings and locate the cluster, we next investigate their structural organization.
% Specifically, we start with high similarity clusters to determine whether certain token groups encode repetitive or input-independent patterns—potentially corresponding to redundant or sink-like representations.

\section{Image-Agnostic Cluster: Attention Sink}
Attention sinks are known to be largely input-independent tokens that attract disproportionate attention despite carrying limited semantic content~\cite{massiveactivations,attentionsink_emprical,attentionsink}. Motivated by this property, we question whether the persistent, high-similarity visual clusters identified earlier correspond to such tokens. Specifically, we analyze two types of visual sinks: (1) \emph{ViT sinks}, which originate from the CLIP vision encoder and are often believed to encode global, image-level context~\cite{ViTRegs,ViTLLMSink}; and (2) \emph{LLM sinks}, which emerge within the language backbone and consistently attend to same regions regardless of the textual input~\cite{visualattentionsink,fan2025mathcalvisimathcalprunerdecodingdiscontinuouscrossmodal}.

% The cross-image high-homogeneity cluster exhibits extreme input-independence, with tokens remaining nearly identical across images. This behavior is the hallmark of "attention sinks"—input-agnostic tokens that serve a structural, non-semantic function by attracting disproportionate attention~\cite{massiveactivations,attentionsink_emprical,attentionsink}. Motivated by this property, we question whether the persistent, high-similarity visual clusters identified earlier correspond to such tokens. Specifically, we analyze two types of visual sinks: (1) \emph{ViT sinks}, which originate from the CLIP vision encoder and are often believed to encode global, image-level context~\cite{ViTRegs,ViTLLMSink}; and (2) \emph{LLM sinks}, which emerge within the language backbone and consistently attend to same regions regardless of the textual input~\cite{visualattentionsink,fan2025mathcalvisimathcalprunerdecodingdiscontinuouscrossmodal}.

\subsection{ViT Attention Sinks}
\vspace{-3pt}
As the visual embeddings originate from the ViT, we intuitively suspect that these clusters may reflect ViT sinks. 

% \paragraph{High-Norm ViT Sinks Align with Redundant Clusters.}
ViT sinks (denoted as $\mathcal{I}_{S_{\text{ViT}}}$) are commonly defined by their abnormally high activation norms. Existing works~\citep{ViTRegs,ViTRegs2,neo2025interpretingvisualinformationprocessing} identify these sinks by setting a norm threshold on the final-layer hidden states of ViT models.  Following \citet{ViTRegs2}, we mark tokens as ViT sinks if their final-layer $L_2$ norm exceeds $\tau=75$ in the last layer of the CLIP. For each image, we then compute the centroid of its ViT sinks, $\mathbf{C}^{S_{\text{ViT}}}$, and measure cross-image similarity between these centroids. Across 1{,}000 randomly sampled images, all $\mathbf{C}^{S_{\text{ViT}}}$ achieve cross-image cosine similarity above $0.99$ with variance below $1\times10^{-5}$, indicating that ViT sinks are nearly identical across inputs. Moreover, these high-norm tokens fall into the same cluster that exhibits the highest cross-image similarity from~\cref{fig:02_cluster}, linking ViT sinks directly to the redundant, input-invariant visual clusters observed earlier. This strong overlap suggests that ViT sinks are consistent across inputs, and while prior works often interpret them as register tokens encoding holistic global information~\cite{ViTRegs,ViTRegs2,liu2025hiprunetrainingfreevisualtoken,ViTLLMSink}, our findings indicate that \emph{they largely reflect input-independent redundancy rather than meaningful image-specific semantics}.

\subsection{LLM Attention Sinks}
\vspace{-3pt}
% \paragraph{Propagation Gap Between ViT and LLM.} 
While ViT sinks exhibit high activation norms in the vision encoder, we observe that these tokens do not directly translate into high attention values within the language backbone. In the LLMs, the model largely disregards $\mathcal{I}_{S_{\text{ViT}}}$. This attention shift indicates that visual  attention sinks in LLMs arise from distinct internal dynamics rather than being inherited from the vision encoder. 

% \paragraph{LLM Sinks Align with Redundant Clusters.} 
\citet{visualattentionsink} observed that visual attention sinks activate the same specific channels as the textual sink token $\langle\text{bos}\rangle$, and identified them by applying a threshold $\tau=20$ on the ratio between sink-channel magnitudes and the RMS norm of the hidden state $\phi(x)=\max|\boldsymbol{x}[d_{\text{sink}}]/\sqrt{\frac{1}{D}\sum^D_{d=1}\boldsymbol{x}[d]^2}|$ (\eg $\mathcal{D}_{\text{sink}}= \{1415, 2533\}$ for MLLM with a LLaMA-2 7B backbone). Although $\mathcal{I}_{S_{LLM}}$ begins to exhibit sink-like behavior only after layer 2, similar to how LLMs use punctuation tokens as sink tokens~\cite{massiveactivations,attentionsink_emprical}, these tokens already display high cross-image similarity ($>0.95$) at the embedding stage, and variance below $1\times10^{-4}$. This suggests that \emph{the model is trained to select a small, consistent subset of image tokens as visual sinks from the very beginning}.

\subsection{Sink Clusters Analysis}
\label{subsec:43}
\vspace{-3pt}
% Given both high in-cluster token similarity and high cross-image cluster similarity, we perform cluster labeling on $\mathcal{I}_{S_{\text{ViT}}}$ and $\mathcal{I}_{S_{\text{LLM}}}$ following the procedure described in \cref{subsec:32_NER4_clu_lab}. 
To facilitate clearer and more interpretable cluster-wise analysis, we employ \emph{EmbedLens} to assign textual references to each cluster described in \cref{subsec:32_embedlens_application}, allowing us to directly track sink clusters across layers. As shown by the blue-highlighted patches in~\cref{fig:21_tSNE_semantics}, $\mathcal{I}_{S_{\text{ViT}}}$ shows highest similarity to the token embedding ``igh'' ($\hat{t}_c{=}1141$), while $\mathcal{I}_{S_{\text{LLM}}}$ aligns most closely with ``jav'' ($\hat{t}_c{=}26673$). These textual anchors provide an intuitive way to locate across layers.

\paragraph{Analysis \#1: Negligible Contribution.}
Using the reference token IDs obtained via \emph{EmbedLens}, we identify and prune both $\mathcal{I}_{S_{\text{ViT}}}$ and $\mathcal{I}_{S_{\text{LLM}}}$ clusters at model input. As shown in~\cref{tab:sink}, the removal of these sink clusters leads to no degradation in model performance, indicating that they contribute little to downstream reasoning. While prior studies have suggested that CLIP sinks may encode global image information, our evidence from pruning and high cross-image similarity both imply that these tokens function as redundant placeholders rather than meaningful carriers of image-level semantics. Additionally, we find that the reason why the performance stays stable is that attention previously assigned to $\mathcal{I}_{S_{\text{LLM}}}$ is reallocated to textual sink tokens in system prompts after pruning.

\begin{figure}[h]
    \centering
    \includegraphics[width=1\linewidth]{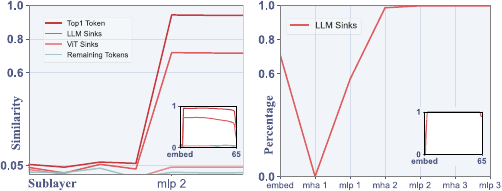}
    \vspace{-20pt}
    \caption{\small \textbf{Formation (Main) and persistence (Insets) of LLM sink tokens.} 
    (Left) Cosine similarity between $\langle\mathrm{bos}\rangle$, ViT sinks, and LLM sinks across sublayers. (Right) Layer-wise tracking of the top-5 most similar tokens to $\langle\mathrm{bos}\rangle$.
    }
    \label{fig:04_sinks_changes}
\end{figure}

\paragraph{Analysis \#2: Layer 2 as Sinks Aligner.}
Given the similar behavior between LLM visual sinks ($\mathcal{I}_{S_{\text{LLM}}}$) and the BOS token $\langle\mathrm{bos}\rangle$, we directly compare their cosine similarity across sublayers. As shown in~\cref{fig:04_sinks_changes}(left), $\mathcal{I}_{S_{\text{LLM}}}$ sharply aligns with $\langle\mathrm{bos}\rangle$ after the second MLP block, indicating that \emph{MLP-2 serves as the sink aligner}, projecting $\mathcal{I}_{S_{\text{LLM}}}$ into an almost identical direction as the BOS representation. 

To verify whether these sink tokens remain stable once formed, we track their evolution layer by layer. Specifically, we select the top-5 most similar tokens to $\langle\mathrm{bos}\rangle$ at MLP-2 and monitor their rank positions in subsequent layers. As shown in~\cref{fig:04_sinks_changes}(right), these tokens persist among the top similarities after MLP-2, yet correlations begin to emerge as early as MLP-1 and the second attention block (MHA-2). This suggests that the representational collapse is progressively shaped through early sublayer interactions, but becomes explicit after MLP-2. We conducted four sublayer skipping experiments (\cref{tab:sink}) and found the performance is stable overall, indicating that the model is generally robust to different forms of sink-token manipulation. Additionally, we find that skipping MLP-2 specifically prevents the semantic alignment between $\mathcal{I}_{S_{\text{LLM}}}$ and $\langle\mathrm{bos}\rangle$; detailed results are provided in the appendix.

\begin{table}[h]
\footnotesize
\caption{%
\textbf{Impact of sink pruning and sublayer skipping on downstream performance.}%
\protect\footnotemark[2] %
``$\notin \mathcal{I}_{S_{\text{LLM}}}$--FF\textsuperscript{2}'' denotes skipping MLP-2 for all \emph{non-sink} visual tokens.}
% ...
\vspace{-10pt}
\begin{tabular}{l|cccc|c}
\toprule
\textbf{Method} & General & OCR & CV Centric & Hallu. & \textbf{Avg.} \\
\midrule
LLaVA-v1.5 7B                     & 58.7 & 36.9 & 54.0 & 61.1 & 52.7 \\

\rowcolor{gray!15}
\multicolumn{6}{c}{\emph{Sinks Pruning}} \\
% \midrule
$- \mathcal{I}_{S_{\text{LLM}}}$  & 58.6 & \textbf{37.0} & \textbf{54.4} & \textbf{61.3} & 52.7 \\
$- \mathcal{I}_{S_{\text{ViT}}}$  & 58.7 & \textbf{37.2} & 53.8 & 61.1 & \textbf{52.8} \\
$- \mathcal{I}_{S}$               & 58.6 & \textbf{37.2} & \textbf{54.1} & \textbf{61.3} & \textbf{52.8} \\

\rowcolor{gray!15}
\multicolumn{6}{c}{\emph{Sublayer Skipping}} \\

$\mathcal{I}_{S_{\text{LLM}}}-$FF\textsuperscript{2}        & 58.7 & \textbf{37.2} & 54.0 & 61.0 & \textbf{52.8} \\
$\mathcal{I}_{S_{\text{LLM}}}-$FF\textsuperscript{1},Att\textsuperscript{2}   & \textbf{58.9} & \textbf{37.1} & 53.9 & \textbf{61.2} & \textbf{52.8} \\
$\mathcal{I}_{S_{\text{LLM}}}-$FF\textsuperscript{1/2},Att\textsuperscript{2} & \textbf{58.9} & \textbf{37.1} & \textbf{54.3} & \textbf{61.2} & \textbf{52.9} \\
$\notin \mathcal{I}_{S_{\text{LLM}}}-$FF\textsuperscript{2} & \textbf{59.0} & \textbf{37.1} & \textbf{54.3} & 60.8 & \textbf{52.8} \\

\bottomrule
\end{tabular}

\label{tab:sink}
\end{table}

\footnotetext[2]{
The General VQA score averages results from GQA~\cite{benchmark:GQA}, MME~\cite{benchmark:mme}, MMBench$^{\text{en}}_{\text{dev}}$~\cite{benchmark:mmbench}, and MMStar~\cite{benchmark:mmstar}. OCR performance averages TextVQA$_{\text{val}}$~\cite{benchmark:textvqa}, OCRBench~\cite{benchmark:ocrbench}, and DocVQA~\cite{benchmark:docvqa}. The CV-centric ability is measured by RefCOCO+ REC val, and RefCOCO REC testA/B~\cite{benchmark:refcoco}. Hallucination is evaluated using POPE~\cite{benchmark:pope} and HallusionBench~\cite{benchmark:HallusionBench}. See the appendix for detailed setups.}

\section{Text-distancing Cluster: Dead Tokens}
\label{sec:deadtokens}
Beyond sink tokens, we further examine the largest and highly repetitive cluster that consistently appears across images, exhibiting strong cross-image similarity yet lack clear semantic alignment.

\subsection{Semantic Inspection}
Across multiple samples, none of these tokens show meaningful or consistent lexical associations at the input stage. When inspected with \emph{EmbedLens}, they frequently correspond to fragmented or non-semantic subword pieces (e.g., punctuation or partial syllables), as shown in the red-highlighted regions of~\cref{fig:21_tSNE_semantics}. This absence of coherent mapping indicates that these repetitive embeddings likely serve no interpretable visual or linguistic function.

% \paragraph{Measuring Contributions and Defining Dead Tokens.}
To quantify their contribution, we prune these repetitive tokens at the input stage. As shown in~\cref{tab:dead_pruning}, removing these tokens even improves task accuracy, confirming that these tokens ($\sim 30\%$) are redundant. We therefore define such tokens as \emph{dead tokens}, as their representations are highly repetitive and semantically void.

\begin{table}[!h]
\footnotesize
\caption{\textbf{Effect of pruning dead-token clusters.}
$\texttt{-Dead}$ removes all dead tokens.
$\texttt{-Remaining}$ randomly removes the same number of dead tokens sampled from the remaining tokens.}
\vspace{-10pt}
\begin{tabular}{l|cccc|c}
\toprule
\textbf{Method} & General & OCR & CV centric & Hallu. & \textbf{Avg.} \\
\midrule
LLaVA-v1.5 7B          & 58.7 & 36.9 & 54.0 & 61.1 & 52.7 \\
\midrule
$\texttt{-Dead}$       & 58.7 & \textbf{37.1} & \textbf{57.7} & \textbf{61.2} & \textbf{53.7} \\
$\texttt{-Remaining}$  & 57.2 & 34.3 & 48.5 & 60.2 & 50.1 \\
\bottomrule
\end{tabular}
\label{tab:dead_pruning}
\end{table}

\subsection{Dead Clusters Analysis}
\label{subsec:dead_analysis}
After identifying repetitive clusters with high cross-image similarity and no semantic grounding, we further analyze when these \emph{dead clusters} form and behave across layers.

\begin{figure}[h]
    \centering
    \includegraphics[width=1\linewidth]{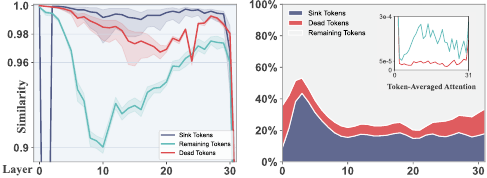}
    \vspace{-20pt}
    \caption{\small \textbf{Dead tokens exhibit minimal representation change and receive limited cross-modal attention.}
\textbf{Left:} Layer-wise cosine similarity within each cluster, showing that dead tokens remain highly self-consistent while other tokens evolve substantially.
\textbf{Right:} Text-to-visual attention distribution across clusters.
Inset: Token-averaged attention comparison.}
    \label{fig:05_dead_evolvement}
\end{figure}

\paragraph{Analysis \#1: Dead Tokens as Inert Residues.}
To examine whether the repetitive clusters evolve semantically or contribute to multimodal reasoning, we analyze their behavior across layers from two perspectives: \emph{(1) in-cluster representation consistency}, and \emph{(2) text-to-visual attention flow}. As shown in~\cref{fig:05_dead_evolvement}, dead tokens maintain almost invariant representations throughout the model, exhibiting minimal semantic drift compared with normal visual tokens. This stability indicates that they are not progressively refined or repurposed by the LLM~\cite{zhao2025skipgpt,men2024shortgpt}. Meanwhile, their cross-modal interaction remains negligible across all layers: the proportion of attention allocated to dead tokens stays below 8\%, and token-averaged attention is also much lower than that of remaining tokens. Together, these findings suggest that \emph{dead tokens neither evolve semantically nor participate in information exchange between modalities}, effectively functioning as static computational residues.

\begin{figure}[h]
    \centering
    \includegraphics[width=1\linewidth]{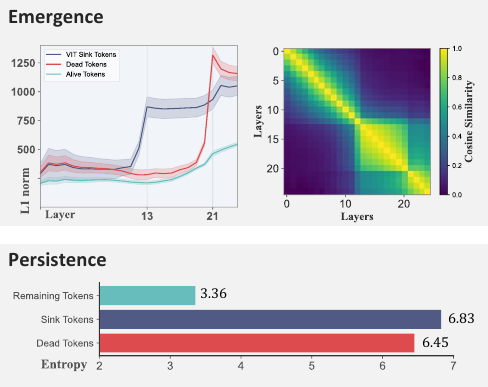}
    \vspace{-20pt}
    \caption{\small\textbf{Emergence and persistence of dead tokens.} 
    (\textbf{Left}) Average L1 norm across ViT layers. 
    (\textbf{Right}) Layer-to-layer cosine similarity in CLIP. 
    (\textbf{Bottom}) Average entropy of cluster semantics at the last layer of LLMs.}
    \label{fig:053_norms}
\end{figure}

\paragraph{Analysis \#2: Emergence and Persistence.}
We first examine the evolution of dead tokens within the CLIP visual encoder. As shown in~\cref{fig:053_norms}, the average embedding norm of dead tokens remains low through the shallow and middle layers but begins to rise sharply around layer 21, coinciding with a clear separation from both normal tokens and ViT sink tokens. This transition marks the point where dead tokens emerge as a distinct group of stable representations. The layer-wise cosine similarity map~\cite{chen2025multimodallanguagemodelsbetter} (right) further confirms this behavior, indicating not only a magnitude increase but also an alignment with a fixed direction in the embedding space. Together, these trends suggest that \emph{dead tokens are formed in the later stages of CLIP’s ViT encoder}, where redundancy accumulates and semantic variation diminishes. Finally, we measure the average entropy of clusters; dead tokens yield significantly higher entropy than remaining visual tokens, confirming the absence of discriminative semantics even at the LLM output stage and showing that this behavior persists throughout the model.

% We further explore whether dead tokens exhibit consistent spatial preferences across images. As shown in Fig.~\ref{fig:053_spatial_distribution}, we categorize input images by aspect ratio—\emph{wide} ($1\times N$), \emph{tall} ($N\times 1$), and \emph{square} ($1\times 1$)—and divide each into a $3\times3$ grid to measure the spatial density of dead tokens.\todo{to quantify} The visualization reveals that dead tokens tend to concentrate along image borders and long edges, regardless of content. This spatial bias suggests that dead tokens originate from redundant boundary patches introduced by uniform patch embedding, which carry limited visual information but persist through the encoding process.

% \begin{figure}[h]
%     \centering
%     \includegraphics[width=1\linewidth]{sec/images/54_image_sizes.pdf}
%     \caption{\small \textbf{Spatial distributions of dead tokens.} 
%     Dead tokens cluster near image borders and long edges across different aspect ratios, indicating redundancy induced by patch embedding rather than semantic relevance. \todo{resolution.}}
%     \label{fig:053_spatial_distribution}
% \end{figure}
\section{Text-proximity Cluster: Alive Tokens}
After isolating sink and dead tokens, we define the remaining subset of visual tokens as \emph{alive tokens}—those that actively participate in multimodal reasoning and contribute meaningful information to the model’s final output. 

\subsection{Informative Tokens}

\paragraph{Semantic Grounding.} As shown in~\cref{fig:21_tSNE_semantics}, t-SNE visualization~\cite{tSNE} of the joint embedding space reveals that alive tokens cluster closely around textual centroids centroids, in contrast to dead tokens that form isolated groups with little semantic alignment. This proximity suggests that \emph{alive tokens are better aligned with the model’s linguistic space}.

\paragraph{Functional Importance.} To validate their contribution, we randomly prune the same number of alive tokens as dead tokens, and observe a noticeable performance drop (\cref{tab:dead_pruning}). This confirms that \emph{alive tokens carry critical object-specific information and serve as the primary bridge connecting visual semantics to textual reasoning}.

\subsection{Alive Cluster Analysis}
\label{subsec:alive_analysis}

\begin{figure}[h]
    \centering
    \includegraphics[width=1\linewidth]{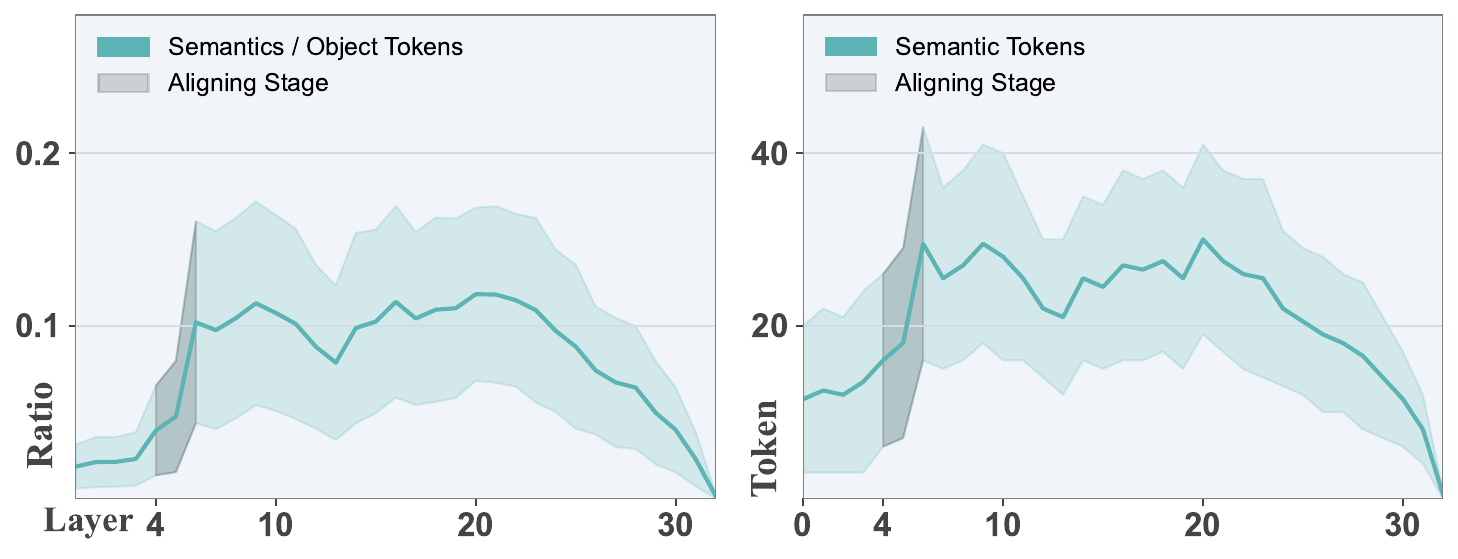}
    \vspace{-20pt}
    \caption{\small \textbf{Semantic sparsity of visual tokens across layers.}
    }
    \label{fig:06_sprsity_of_semantics}
\end{figure}

\paragraph{Analysis \#1: Sparse Distribution of Object Semantics.} Having established that alive tokens encode object-specific information, we next examine how densely such semantics are distributed across visual tokens. Using the MS COCO 2017 dataset with annotated bounding boxes and entity labels~\cite{coco}, we define tokens located within the bounding boxes as \emph{object tokens}. We then apply \emph{EmbedLens} to each layer to identify how many of these object tokens retrieve their annotated object labels. As shown in~\cref{fig:06_sprsity_of_semantics}, only around 2\% of object tokens exhibit correct object semantics at the input, but this proportion quickly rises through a \emph{semantic-alignment stage} and stabilizes around 10\% after the 7th layer. When considering all visual tokens, the fraction containing any grounded semantics remains below 7\% at its peak. These results indicate that semantic grounding in visual embeddings is highly \emph{sparse}, offering a explanation for why even random visual-token pruning or image downsampling can be highly effective on many object-level VQA tasks ~\cite{peng2025visualinputcompressedvisual,wen2025tokenpruningmultimodallarge,llavolta,liao2025usingrightbenchmarkevaluation}: only a small fraction of tokens carry meaningful object semantics.

\begin{figure}[h]
    \centering
    \includegraphics[width=1\linewidth]{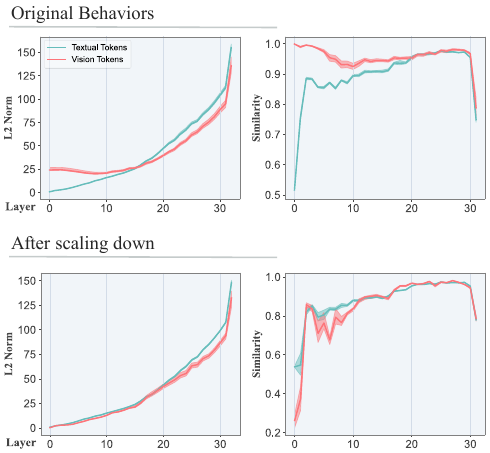}
    \vspace{-20pt}
    \caption{\small \textbf{Layer-wise alignment dynamics and effect of norm scaling.}  }
    \label{fig:06_norm}
\end{figure}

\paragraph{Analysis \#2: High Norms Suppress Early-Layer Processing.} To quantify the visual transformation within the LLM backbone, we compare the evolution of \emph{visual} and \emph{textual} token norms and layer-wise cosine similarities. As shown in~\cref{fig:06_norm}(top), visual tokens consistently exhibit substantially higher $L_2$ norms than textual tokens. Given that high-norm representations dilute the relative effect of sublayer updates on the residual stream~\cite{norm_chen2025pruneandcomp,norm_csordás2025languagemodelsusedepth}, we hypothesize that shallow layers exert limited influence on visual embeddings. To test this, we scaled down the projected visual embeddings by a factor of $0.01$ before feeding them into the LLM. As shown in~\cref{fig:06_norm}(bottom), the reduced norm reintroduces active transformations resembling textual tokens, but leads to degraded visual understanding. This finding suggests that the projector deliberately amplifies visual norms to \emph{bypass redundant shallow-layer processing}, directly aligning visual tokens to mid-layer representations that interface more effectively with textual reasoning. 
\section{How Semantics Are Decoded and Refined}
With the semantic structure of visual tokens established, we now ask how these tokens are decoded and transformed by MLLMs. Visual tokens often encode multiple semantic cues, yet their representations change only minimally through the language backbone. This contrast prompts two questions: How well can a model extract multiple semantic attributes from a single patch, and how does its structure shape visual processing?

\subsection{Decoding Multiple Trajectories in MLLMs}
\label{subsec:bench}

As shown in~\cref{fig:21_tSNE_semantics}, a single visual token often encodes multiple visual cues (\eg object identity, color, and counting). While recent studies suggest that LLMs may follow only a dominant semantic trajectory within such soft tokens~\cite{wu2025softtoken}, to examine whether a single patch embedding with multiple semantic attributes can be effectively decoded, we construct a lightweight diagnostic benchmark.

Each sample contains a single object or character rendered at varying scales to ensure it precisely fits within one visual patch. For each image, we design three question types aligned with distinct semantic axes identified earlier: object recognition, color identification, and counting. The dataset consists of 70 images across three groups: object (30), OCR (20), and OCR with background color different from the padding color, designed to test whether models rely on dominant patch color rather than object-grounded color. 
% Specifically, the original image is resized to a square, scaled by a factor proportional to the grid size, and then resized to a fixed resolution of 448×448. 
\begin{figure}[h]
    \centering
    \includegraphics[width=1\linewidth]{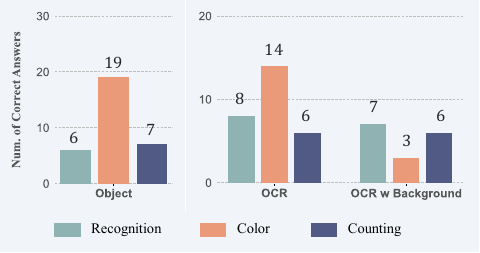}
    \vspace{-20pt}
    \caption{\small \textbf{Results on our multi-trajectory benchmark.}}
    \label{fig:71_bench}
\end{figure}

Based on the evaluation results in~\cref{fig:71_bench}, we summarize three major insights:
\begin{enumerate}
    \item \textbf{Multi-Semantic Compositionality.} Among the six correctly recognized objects, five also correctly answered both color and counting questions, indicating that MLLMs not only encode multiple semantic trajectories within a single patch but also \emph{demonstrate the ability to reason over them}\footnote{Notably, correct color predictions cannot be explained by object priors (e.g., “yellow bus”, or “green bananas”).}.
    \item \textbf{Tokenization Limits Fine-Grained OCR.} For OCR-based samples, accuracy drops substantially when the rendered text is tokenized into multiple subwords (e.g., “tT” or “gb”). In contrast, contiguous strings such as “Fu,” “Ty,” or even four-letter words like “Jack” are correctly recognized. 
     \item \textbf{Contextual Color Bias.} Color recognition accuracy declines sharply when object and background colors differ (right panel in~\cref{fig:71_bench}). Incorrect predictions tend to follow the surrounding background, suggesting that VLMs often associate color with dominant regional statistics rather than object-grounded features. Detailed qualitative analysis is provided in the Appendix. 
     
\end{enumerate}

\subsection{Decoupling Visual Transformation in MLLMs}
\label{subsec:visual_processing}

Guided by earlier evidence that (i) visual tokens change minimally across shallow layers, and (ii) different sublayers play asymmetric roles (\cref{subsec:dead_analysis}), we design two complementary experiments: \textbf{(1) Shallow-Layer Skipping.} Since early-layer transformations contribute little to visual-token refinement, we input visual tokens only up to a chosen layer and bypass all preceding visual processing. This allows us to assess how much shallow computation can be removed without impairing multimodal reasoning. \textbf{(2) Layer Decoupling.} Inspired by the prominent role of sublayer in sink formation, we selectively disable visual-only sublayers by zeroing their visual outputs (\texttt{vMHA} or \texttt{vFFN}), ensuring they do not alter the visual residual stream. This isolates the functional contribution of each sublayer while preserving cross-modal information flow.

\begin{table}[!h]
\footnotesize
\caption{\small \textbf{Effect of layer decoupling and shallow-layer skipping on multimodal reasoning.} \textsuperscript{$\dagger$} denotes training-based method. }
\vspace{-10pt}
\begin{tabular}{l|cccc|c}
\toprule
\textbf{Method} & General & OCR & CV Centric & Hallu. & \textbf{Avg.} \\
\midrule
LLaVA-v1.5 7B               & 58.7 & 36.9 & 54.0 & 61.1 & 52.7 \\
\rowcolor{gray!15} \multicolumn{6}{c}{\emph{Layer Decoupling}} \\
% $-\texttt{vMHA}$            & ---- & ---- & ---- & ---- & ---- \\
$-\texttt{vMHA}^{\dagger}$  & 58.7 & 35.4 & 37.2 & 62.0 & 48.3 \\
% $-\texttt{vFFN}$            & ---- & ---- & ---- & ---- & ---- \\
$-\texttt{vFFN}^{\dagger}$  & 57.8 & 35.4 & 50.6 & 61.7 & 51.4 \\
\rowcolor{gray!15} \multicolumn{6}{c}{\emph{Shallow Layers Skipping}} \\
$\texttt{L}_{\texttt{in}}\texttt{=4}^{\dagger}$  & 58.7 & 36.6 & 56.4 & 61.2 & 53.2 \\
$\texttt{L}_{\texttt{in}}\texttt{=5}^{\dagger}$  & 58.4 & 36.7 & 56.8 & 61.1 & 53.3 \\
$\texttt{L}_{\texttt{in}}\texttt{=6}^{\dagger}$  & 58.0 & 36.6 & 56.0 & 61.1 & 52.9 \\
$\texttt{L}_{\texttt{in}}\texttt{=8}^{\dagger}$  & 57.9 & 35.9 & 52.7 & 61.1 & 51.9 \\
$\texttt{L}_{\texttt{in}}\texttt{=10}^{\dagger}$ & 57.0 & 36.0 & 52.0 & 60.1 & 51.3 \\

\bottomrule
\end{tabular}

\label{tab:no_vis}
\end{table}

\paragraph{Layer Decoupling} 
Results in~\cref{tab:no_vis} show that the necessity of visual processing for MHA and FFN is highly task-dependent. For hallucination-related benchmarks, disabling either processing even improves performance, suggesting that redundant visual self-updates may amplify noise. General reasoning tasks remain largely unaffected; however, \texttt{vFFN} appears more essential than \texttt{vMHA}, indicating that feed-forward modules may inject task-relevant knowledge, while most intra-visual interactions have already been captured by the CLIP encoder. CV-centric benchmarks show clear degradation, particularly when \texttt{vMHA} is removed, highlighting that spatially grounded reasoning still relies on token–token attention within the visual stream, even though it contributes little to high-level semantic fusion. OCR tasks, meanwhile, rely on both \texttt{vMHA} and \texttt{vFFN}, suggesting that both fine-grained spatial reasoning and knowledge refinement are required for accurate text recognition.

\paragraph{Shallow Layers Skipping} The shallow-layer skipping results produce three key insights:
\begin{itemize}
    \item \textbf{Minimal need for shallow visual computation.} Echoing~\cref{subsec:alive_analysis}, skipping the first 6 visual layers preserves or slightly improves general performance, confirming that early layers contribute little to refining visual tokens.
    \item \textbf{Shallow layers disrupt spatial structure.} Accuracy on CV-centric tasks improve notably skipping the first 5–7 layers, suggesting that early visual processing may introduce noise that obscures spatial cues.
    \item \textbf{Projection does not increase semantic grounding.} Late-entry models begin with nearly the same semantic-object ratio as the vanilla model, but the semantic-alignment stage becomes more evenly distributed across following layers. (See appendix for details).
\end{itemize}

\section{Related Work}

\paragraph{Multi-modal Large Language Models.} Recent multimodal LLMs integrate modality-specific encoders with large language models through lightweight projection, enabling effective reasoning across diverse modalities~\cite{llava,Qwen2.5-VL,zhu2025internvl3,qiu2025few,tong2025llm,sun2025llaso,qiu2026rethinking,Wu2026survey}. While early systems relied heavily on cross-attention or specialized fusion modules~\cite{blip2,flamingo}, recent models increasingly adopt simple projection-based designs that map visual features into the LLM token space~\cite{liu2026vicaefficientmultimodalllms}. Despite strong performance, the semantic behavior of visual tokens across layers remains unclear, and we provide a fine-grained analysis of their representation and functional role.

\paragraph{Vision-Text Alignment in VLMs.} Recent work shows that independently trained vision and language models converge toward similar representational structures. The “Platonic Representation Hypothesis”~\cite{huh2024platonicrepresentationhypothesis} formalizes this phenomenon, suggesting that as model scale grows, vision and language embeddings are aligning across architectures and training paradigms~\cite{maniparambil2024visionlanguageencodersrepresent}, with even text-only LLMs developing implicit visual priors~\cite{han2025learningseeingdemystifyingllm}.
In the multimodal setting, identified by logits lens~\cite{logitlens}, it is commonly believed that visual tokens in VLMs are aligning with the textual embedding space from the middle to late layers~\citep{chen2025multimodal,venhoff2025visualrepresentationsmaplanguage,he2025seeingwordsspeakingpixels,CircuitProbe,neo2025interpretingvisualinformationprocessing,wu2025semantichubhypothesislanguage,wu2026hidivdrop}. In this paper, we provide further evidence that visual embeddings have been partially aligned with the language model’s input embedding space via \emph{EmbedLens}.

\section{Conclusions}
We provide a semantic-centered view of how multimodal LLMs embed, decode, and refine visual information. Our findings uncover a stable tri-partition of visual tokens, expose how these representations are decoded into downstream semantics, and reveal that not all sublayers contribute meaningfully to semantic refinement. These insights outline a coherent mechanism of semantic flow inside MLLMs and offer practical guidance for building more interpretable and efficient multimodal systems.

{
    \small
    \bibliographystyle{ieeenat_fullname}
    \bibliography{main}
}
\clearpage
\setcounter{page}{1}
\setcounter{section}{0}

\renewcommand{\thesection}{\Alph{section}}

\setcounter{figure}{0}
\renewcommand{\thefigure}{\alph{figure}}

\setcounter{table}{0}
\renewcommand{\thetable}{\alph{table}}

\maketitlesupplementary

\section{Evaluation Setups}
\paragraph{General Tasks}
We assess general visual question answering ability on four standard multimodal benchmarks: GQA, MME, MMBench$^{\text{en}}_{\text{dev}}$, and MMStar~\cite{benchmark:GQA,benchmark:mme,benchmark:mmbench,benchmark:mmstar}. GQA contains compositional reasoning questions over real-world images. MME perception is a comprehensive evaluation suite covering 14 perception subtasks. 
MMBench$^{\text{en}}_{\text{dev}}$ is a multiple-choice benchmark with about 3k questions over 20 ability dimensions; MMStar is a vision-indispensable benchmark with 1.5k carefully curated samples that cover 6 core capabilities and 18 axes. The \emph{General VQA} score reported in the main text is the unweighted mean of the normalized scores on these four datasets, the score for MME is divided by 20.

\paragraph{OCR Tasks}
To evaluate text-centric perception, we use TextVQA$_{\text{val}}$, OCRBench, and DocVQA~\cite{benchmark:ocrbench,benchmark:textvqa,benchmark:docvqa}. TextVQA requires reading scene text and answering open-ended questions.OCRBench is a composite OCR benchmark covering text recognition, scene-text VQA, document VQA, key information extraction, and handwritten math expression recognition. DocVQA measures document understanding via question answering on document images. The \emph{OCR} score is the unweighted mean of the normalized TextVQA, OCRBench, and DocVQA scores.

\paragraph{CV Centric Tasks}
We use referring expression comprehension (REC) benchmarks to measure fine-grained, localization-centric vision ability. Specifically, we evaluate on RefCOCO+ REC val and RefCOCO REC testA/B~\cite{benchmark:refcoco}, which localize a target region in a COCO image given a natural-language referring expression. Following common practice, we use the Acc@0.5 metric, which is the
standard detection metric for REC.  The \emph{CV-Centric} score is the average accuracy over the three REC splits (RefCOCO+ val, RefCOCO testA, and RefCOCO testB).

\paragraph{Hallucination}
Hallucination robustness is evaluated on POPE and HallusionBench~\cite{benchmark:pope,benchmark:HallusionBench}. POPE is a polling-based object probing benchmark that tests whether a model correctly judges the existence of objects using yes/no questions under random, popular, and adversarial sampling strategies, we report F1 scores. HallusionBench is an image-context reasoning benchmark designed to disentangle language hallucination and visual illusion, we use DeepSeek-V3.2-Exp Chat~\cite{deepseekai2025deepseekv3technicalreport} as the evaluator and we report the All Accuracy(aAcc) . The overall \emph{Hallucination} score reported in the paper is the mean of the normalized POPE and HallusionBench scores.

\subsection{Detailed Results}

This section provides the full experimental results corresponding to the three summary tables in the main text.
\begin{itemize}
    \item For \cref{tab:sink} (impact of sink pruning and sublayer skipping), detailed metrics are reported in Appendix \cref{apptab:sink}.
    \item For \cref{tab:dead_pruning} (effects of pruning dead tokens), complete results are provided in Appendix \cref{apptab:dead}.
    \item For \cref{tab:no_vis} (training-based visual processing analysis), detailed results appear in Appendix \cref{apptab:trained}.
\end{itemize}

\begin{table*}[t]
\caption{\textbf{Effect of sink pruning and sublayer skipping.} 
Performance remains stable under pruning or skipping operations, indicating weak reliance on visual sink tokens. 
$\notin\mathcal{I}_{S_{\text{LLM}}}$–MLP2 skips MLP-2 only for non-sink visual tokens.}

\begin{tabular}{l|cccc|ccc|ccc|cc|c}
\toprule
\multirow{2}{*}{Method} &  \multicolumn{4}{c|}{General} & \multicolumn{3}{c|}{OCR} & \multicolumn{3}{c|}{CV Centric} & \multicolumn{2}{c|}{Hallucination} & \textbf{Avg.} \\
& \rotatebox{90}{GQA} & \rotatebox{90}{MME\textsuperscript{P}} & \rotatebox{90}{MMStar} & \rotatebox{90}{MMBench} & \rotatebox{90}{VQA\textsubscript{text}} & \rotatebox{90}{OCRbench} & \rotatebox{90}{VQA\textsubscript{Doc}} & \rotatebox{90}{RefCOCO+\textsubscript{val}} & \rotatebox{90}{RefCOCO\textsubscript{testA}} & \rotatebox{90}{RefCOCO\textsubscript{testB}} & \rotatebox{90}{POPE} & \rotatebox{90}{Hallusion} & \rotatebox{90}{Avg.} \\
\midrule
LLaVA-v1.5 7B & 61.9 & 1507 & 33.5 & 64.1 & 58.2 & 31.2 & 21.5 & 50.0 & 64.5 & 47.4 & 85.9 & 36.2 & 52.7 \\

\rowcolor{gray!15} \multicolumn{14}{c}{\emph{Sinks Pruning}} \\
$-\texttt{ViT Sinks}$ & 61.9 & 1502 & 33.2 & 64.5 & 58.2 & 31.5 & 21.9 & 50.4 & 65.1 & 47.6 & 85.9 & 36.3 & 52.8 \\
$-\texttt{LLM Sinks}$ & 62.0 & 1488 & 33.5 & 63.8 & 58.2 & 31.3 & 21.5 & 50.1 & 64.2 & 47.0 & 85.9 & 36.6 & 52.7 \\
$-\texttt{All Sinks}$ & 62.0 & 1499 & 33.1 & 64.3 & 58.2 & 31.4 & 21.9 & 50.5 & 64.5 & 47.4 & 85.9 & 36.6 & 52.8 \\

\rowcolor{gray!15} \multicolumn{14}{c}{\emph{Sublayer Skipping}} \\
$\mathcal{I}_{S_{\text{LLM}}}-$\texttt{MLP2} & 62.0 & 1501 & 32.9 & 64.7 & 58.2 & 31.7 & 21.6 & 50.7 & 64.9 & 47.8 & 85.7 & 36.2 & 52.8    \\
$\mathcal{I}_{S_{\text{LLM}}}-$\texttt{MLP1,ATT2} & 61.9&1517&33.5&64.3&58.2&31.6&21.5&50&64.3&47.4&85.9&36.4  & 52.8  \\
$\mathcal{I}_{S_{\text{LLM}}}-$\texttt{MLP1/2,ATT2} & 62.0&1513&33.4&64.7&58.2&31.5&21.5&50.5&64.6&47.8&85.8&36.6 & 52.8   \\
$\mathcal{I}_{S_{\text{LLM}}}-$\texttt{MLP1/2,ATT2} & 62.0&1513&33.4&64.7&58.2&31.5&21.5&50.5&64.6&47.8&85.8&36.6 & 52.9   \\
$\notin \mathcal{I}_{S_{\text{LLM}}}-$\texttt{MLP2} & 61.9&1525&33.6&64.2&58.2&31.6&21.5&50.4&64.5&48.1&85.7&35.8 & 52.8   \\

\bottomrule
\end{tabular}

\label{apptab:sink}
\end{table*}

\begin{table*}[t]
\caption{\textbf{Effect of pruning dead-token clusters.} 
Pruning dead-token clusters yields a small boost in accuracy, whereas removing the same number of remaining visual tokens causes substantial drops across all benchmarks. This validates that dead tokens are semantically void and redundant, while non-dead tokens carry meaningful information.}
\begin{tabular}{l|cccc|ccc|ccc|cc|c}
\toprule
\multirow{2}{*}{Method} &  \multicolumn{4}{c|}{General} & \multicolumn{3}{c|}{OCR} & \multicolumn{3}{c|}{CV Centric} & \multicolumn{2}{c|}{Hallucination} & \textbf{Avg.} \\
& \rotatebox{90}{GQA} & \rotatebox{90}{MME\textsuperscript{P}} & \rotatebox{90}{MMStar} & \rotatebox{90}{MMBench} & \rotatebox{90}{VQA\textsubscript{text}} & \rotatebox{90}{OCRbench} & \rotatebox{90}{VQA\textsubscript{Doc}} & \rotatebox{90}{RefCOCO+\textsubscript{val}} & \rotatebox{90}{RefCOCO\textsubscript{testA}} & \rotatebox{90}{RefCOCO\textsubscript{testB}} & \rotatebox{90}{POPE} & \rotatebox{90}{Hallusion} & \rotatebox{90}{Avg.} \\
\midrule
LLaVA-v1.5 7B & 61.9 & 1507 & 33.5 & 64.1 & 58.2 & 31.2 & 21.5 & 50.0 & 64.5 & 47.4 & 85.9 & 36.2 & 52.7 \\
\midrule
\texttt{-Dead Tokens} & 61.8& 1495& 34& 64.1& 58.2& 31.6& 21.6& 53.4& 68.4& 51.4& 86& 36.3& 53.7 \\
\texttt{-Alive Tokens} & 60.5& 1475& 32.2& 62.5& 55.4& 29& 18.6& 44.4& 58.3& 42.9& 84.2& 36.2& 50.1 \\
\bottomrule
\end{tabular}
\label{apptab:dead}
\end{table*}

\begin{table*}[t]
\caption{\textbf{Effect of layer decoupling and shallow-layer skipping on multimodal reasoning.} \texttt{vMHA} is crucial for spatially grounded reasoning in CV-centric tasks, while \texttt{vFFN} contributes to knowledge refinement. In contrast, shallow-layer skipping shows minimal degradation, confirming that early visual processing is largely redundant. \textsuperscript{$\dagger$} denotes training-based method.}

\begin{tabular}{l|cccc|ccc|ccc|cc|c}
\toprule
\multirow{2}{*}{Method} &  \multicolumn{4}{c|}{General} & \multicolumn{3}{c|}{OCR} & \multicolumn{3}{c|}{CV Centric} & \multicolumn{2}{c|}{Hallucination} & \textbf{Avg.} \\
& \rotatebox{90}{GQA} & \rotatebox{90}{MME\textsuperscript{P}} & \rotatebox{90}{MMStar} & \rotatebox{90}{MMBench} & \rotatebox{90}{VQA\textsubscript{text}} & \rotatebox{90}{OCRbench} & \rotatebox{90}{VQA\textsubscript{Doc}} & \rotatebox{90}{RefCOCO+\textsubscript{val}} & \rotatebox{90}{RefCOCO\textsubscript{testA}} & \rotatebox{90}{RefCOCO\textsubscript{testB}} & \rotatebox{90}{POPE} & \rotatebox{90}{Hallusion} & \rotatebox{90}{Avg.} \\
\midrule
LLaVA-v1.5 7B & 61.9 & 1507 & 33.5 & 64.1 & 58.2 & 31.2 & 21.5 & 50.0 & 64.5 & 47.4 & 85.9 & 36.2 & 52.7 \\

\rowcolor{gray!15} \multicolumn{14}{c}{\emph{Layer Decoupling}} \\
-\texttt{vMHA}$^{\dagger}$ & 61.9 & 1472 &32.1&63.5&56.9&30.3&19&45.1&62.3&44.5&86.6&36.7&51.4 \\
-\texttt{vFFN}$^{\dagger}$ &62.2& 1507& 33.7& 63.5& 56.3& 31.2& 18.7& 33.5& 46& 32.1& 86.6& 37.3& 48.3 \\

\rowcolor{gray!15} \multicolumn{14}{c}{\emph{Shallow Layers Skipping}} \\
$\texttt{L}_{\texttt{in}}\texttt{=4}^{\dagger}$ & 62.4& 1444& 34.0& 66.0& 55.9& 32.0& 21.8& 51.2& 67.9& 50.0& 85.6& 36.8& 53.2 \\
$\texttt{L}_{\texttt{in}}\texttt{=5}^{\dagger}$ & 62.4& 1442& 33.3& 65.8& 56.5& 31.9& 21.6& 51.9& 68.0& 50.6& 85.8& 36.4& 53.3 \\
$\texttt{L}_{\texttt{in}}\texttt{=6}^{\dagger}$ & 62.2& 1413& 33.1& 66.2& 57.3& 31.3& 21.2& 51.4& 67.3& 49.3& 85.4& 36.8& 52.9 \\
$\texttt{L}_{\texttt{in}}\texttt{=8}^{\dagger}$ & 61.5& 1440& 32.7& 65.4& 56.1& 31.0& 20.5& 48.0& 64.0& 46.1& 85.4& 36.7& 51.9 \\
$\texttt{L}_{\texttt{in}}\texttt{=10}^{\dagger}$ & 61.9& 1391& 33.3& 63.1& 57.6& 30.5& 20.0& 46.7& 63.8& 45.5& 85.9& 34.2& 51.3 \\

\bottomrule
\end{tabular}

\label{apptab:trained}
\end{table*}

\section{Implementations of Clustering Analysis}
\subsection{Intra-image Visual Embeddings Clustering}
\paragraph{Anchor-based Clustering.}
We perform an anchor-based clustering over normalized post-projection visual embeddings. For each anchor $\mathbf{v}_a$, a cluster $\mathcal{C}_a$ is formed by grouping embeddings whose cosine similarity with the anchor exceeds a threshold $\tau$:
\[
\mathcal{C}_a = \{\, j \mid \frac{\langle \mathbf{v}_a, \mathbf{v}_j \rangle}{\|\mathbf{v}_a\|_2 \|\mathbf{v}_j\|_2} \ge \tau \,\}.
\]

\paragraph{Dominance of a Single Visual Cluster.} 
With the similarity threshold set to $\tau = 0.9$, we rank clusters by their token counts for each image.  
~\cref{fig:02_cluster}(left) shows the top five clusters ranked by size. The largest cluster, denoted as $\mathcal{C}_0$, consistently contains far more tokens than all remaining clusters combined. This indicates that, for each image, around 30\% of post-projection visual embeddings encode highly repetitive information. We next investigate whether this redundancy is image-specific or reflects image-agnostic representational patterns shared across inputs.

\subsection{Cross-image Cluster Similarity}

To evaluate the consistency of visual clusters across different images, we compute the maximum cross-image similarity between cluster centroids. For each input image, we perform the same anchor-based clustering, and normalize the resulting cluster centroids. Given two sets of normalized cluster centroids $\mathbf{C}^{(1)}$ and $\mathbf{C}^{(2)}$, their cross-image similarity is defined as:
\[
S_{ij} = \mathbf{C}^{(1)}_i \cdot \mathbf{C}^{(2)}_j,
\quad
\text{and} \quad
s_i = \max_j S_{ij},
\]
where $s_i$ measures the highest alignment of cluster $i$ from one image to any cluster in another. Averaging $s_i$ over multiple image pairs provides a compact measure of cluster stability and semantic coherence across visual instances. 

\paragraph{Persistence of Clusters Across Images.} 
As shown in~\cref{fig:02_cluster}(right) for the top six most similar clusters, our cross-image analysis reveals that \emph{many visual tokens not only encode identical information within their own image but also occupy nearly the same regions in the representation space across different images}. For example, the largest cluster $\mathcal{C}_0$ of each image remains remarkably stable, with an average cross-image similarity exceeding $\mathbf{S}_{\mathbf{C}} > 0.98$ and variance below $10^{-4}$. Several smaller clusters also exhibit strong cross-image consistency, suggesting shared latent semantics or structural priors. These findings highlight the need for further semantic validation of such persistent tokens, as their invariance may stem from modality-specific artifacts rather than meaningful visual cues.

\subsection{Impacts of $\tau$}

\begin{figure}[h]
    \centering
    \includegraphics[width=1\linewidth]{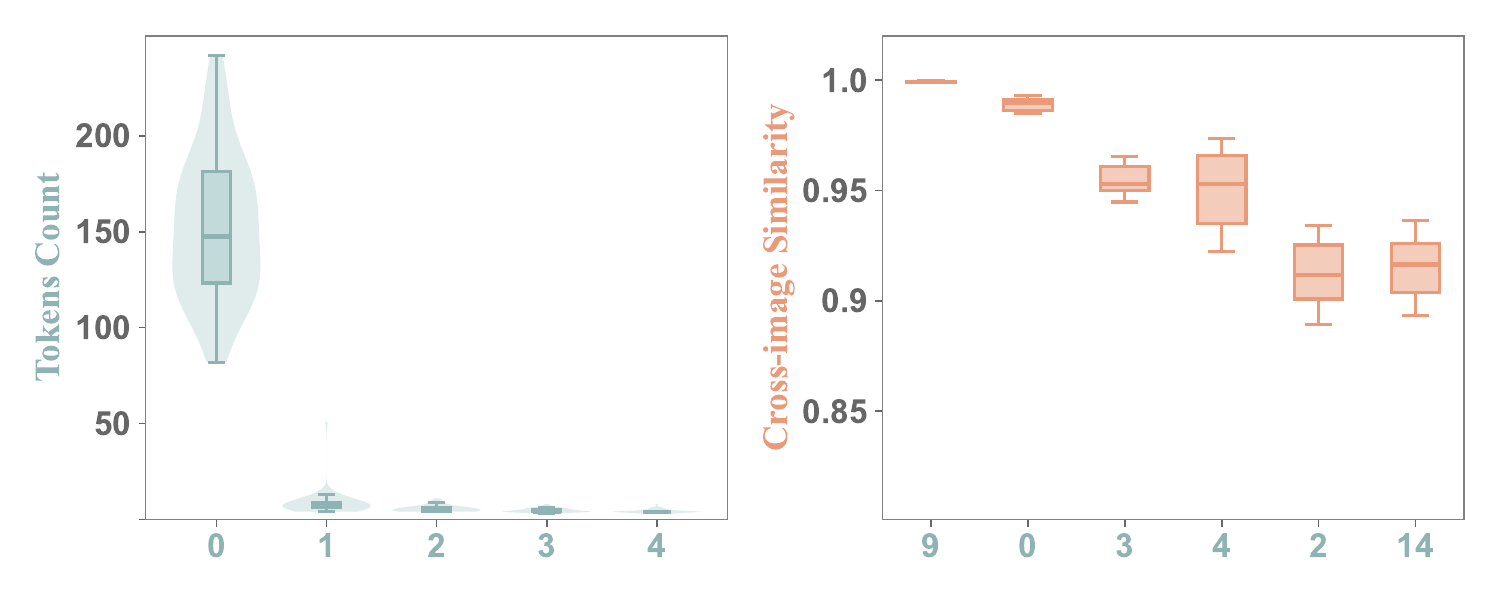}
    \caption{\textbf{Clusters Distribution when $\tau=0.95$}. (Left) Number of tokens in the top-5 clusters. (Right) Cluster similarity cross images. The cluster index is ranked by the number of tokens.}
    \label{appfig:B_tau_95}
\end{figure}

\begin{figure}[h]
    \centering
    \includegraphics[width=1\linewidth]{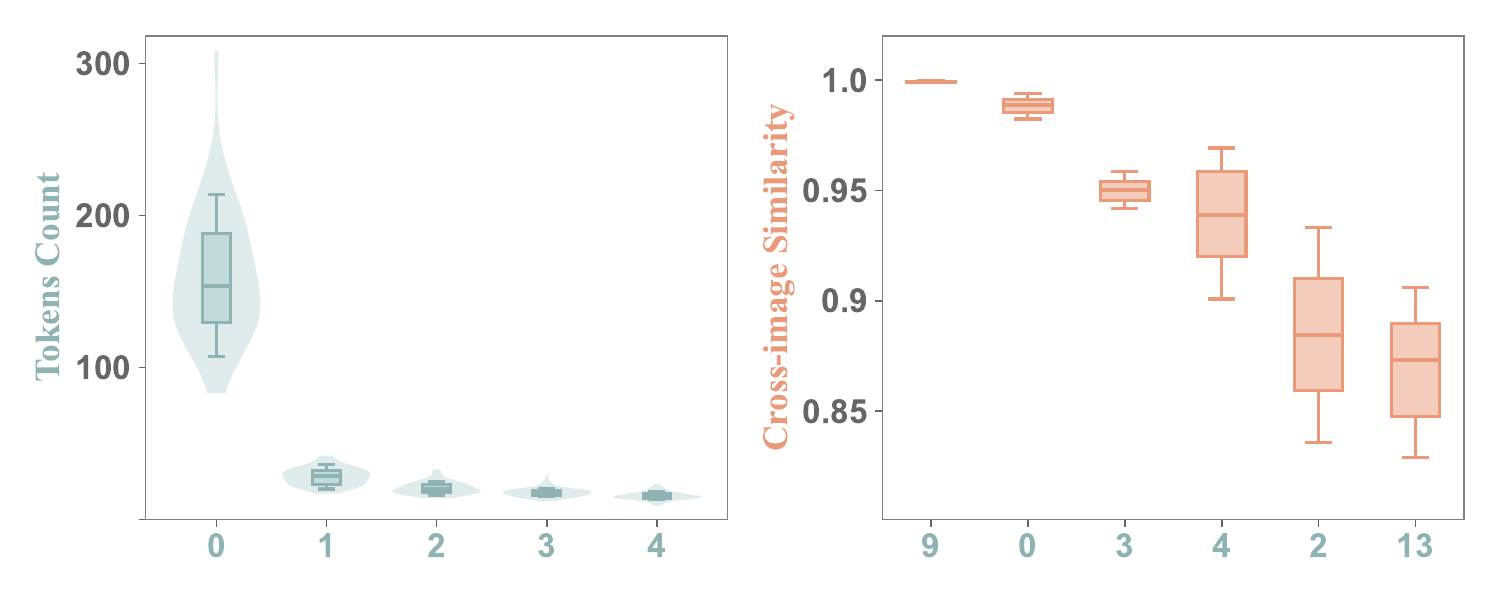}
    \caption{\textbf{Clusters Distribution when $\tau=0.8$}. (Left) Number of tokens in the top-5 clusters. (Right) Cluster similarity cross images. The cluster index is ranked by the number of tokens.}
    \label{appfig:B_tau_80}
\end{figure}

We also examine the effect of the clustering threshold $\tau$. As shown in \cref{appfig:B_tau_95} and \cref{appfig:B_tau_80}, the overall cluster distribution remains consistent across $\tau = 0.95$, $0.9$, and $0.8$. This indicates that our clustering results are robust and largely insensitive to the choice of $\tau$.

\section{Clustering Analysis on Other Models}
\begin{figure}[h]
    \centering
    \includegraphics[width=1\linewidth]{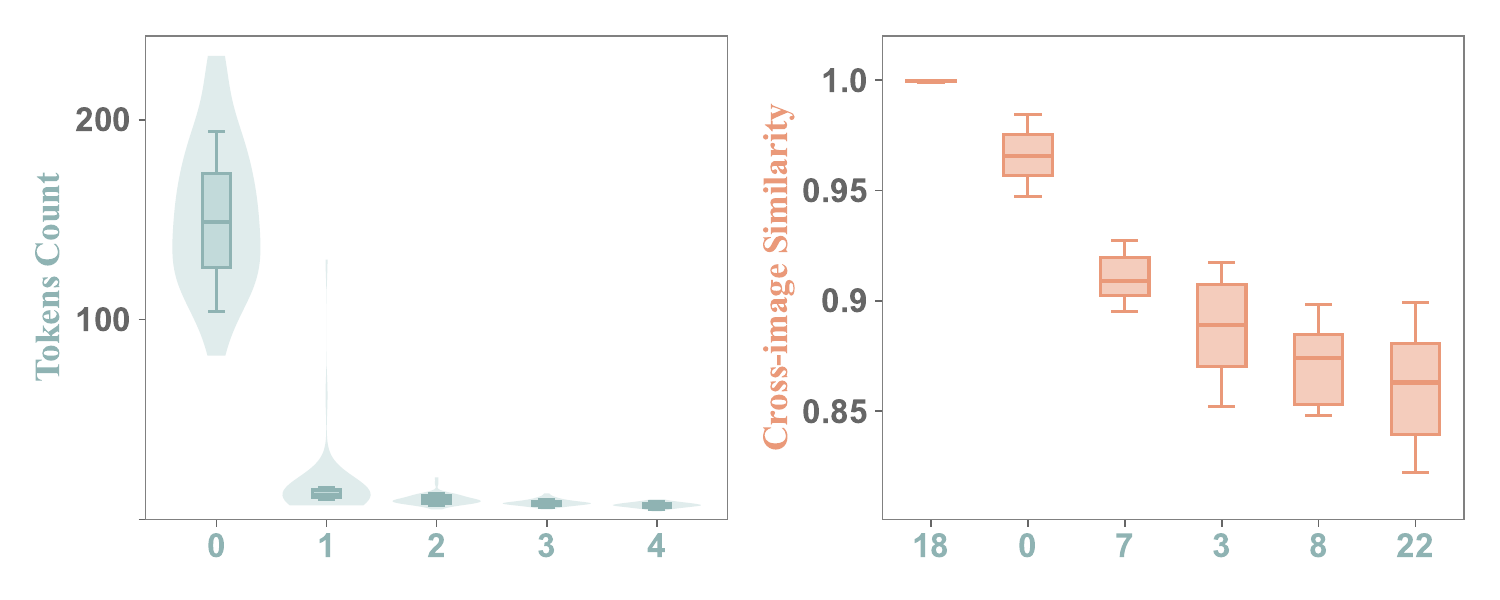}
    \caption{\textbf{Clusters Distribution of LLaVA-v1.5 13B}. (Left) Number of tokens in the top-5 clusters. (Right) Cluster similarity cross images. The cluster index is ranked by the number of tokens.}
    \label{appfig:C_llava_13B}
\end{figure}

Since the projection layer in MLLMs performs only a linear transformation on CLIP ViT features, we hypothesize that the redundant clusters originate primarily from the CLIP visual encoder itself. As shown in \cref{appfig:C_llava_13B}, LLaVA-v1.5-13B exhibits nearly identical clustering behavior to the 7B variant (\cref{fig:02_cluster}), supporting this assumption.

\subsection{Other Vision Encoders}
We evaluate Qwen2.5-VL, InternVL3, and Qwen3-VL, and find that none display the same repetitive clustering pattern. Their visual tokens show notably higher diversity, suggesting that more recent encoders embed richer and less redundant visual information.

\section{Effects of LLM Visual Sinks Manipulation}
\subsection{LLM Sinks Pruning}
As shown in \cref{appfig:D_sinks_manipulation}(left), pruning visual sink tokens at the embedding stage leads to no measurable drop in performance. To understand how the attention previously focused on these tokens is redistributed, we visualize the average attention flow after pruning. The results show that attention originally directed toward visual sinks is reallocated to textual sinks within the system prompt, suggesting that these tokens act primarily as structural placeholders for attention normalization rather than meaningful information carriers.

\subsection{Skipping MLP 2}
\begin{figure}[h]
    \centering
    \includegraphics[width=1\linewidth]{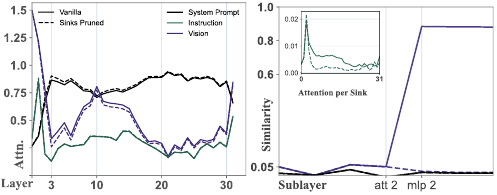}
    \caption{\textbf{Effect of LLM sink manipulation.}
    (Left) Attention redistribution after pruning visual sinks at the input: attention shifts from visual sinks to textual sinks in the system prompt.
    (Right) Skipping MLP-2 weakens alignment between visual sinks and BOS, confirming its role in sink formation, while overall attention patterns remain stable.}
    \label{appfig:D_sinks_manipulation}
\end{figure}

We further test the effect of skipping MLP-2 for sink tokens, the layer identified as the key stage for sink formation. As illustrated in \cref{appfig:D_sinks_manipulation}(right), bypassing MLP-2 substantially reduces the similarity between sink tokens and the $\langle\mathrm{bos}\rangle$ representation, confirming that MLP-2 drives representational collapse toward the $\langle\mathrm{bos}\rangle$ direction. Moreover, the inset plot reveals that skipping MLP-2 also markedly decreases the total attention allocated to visual sinks, suggesting that weakened sink alignment diminishes their ability to attract residual attention across layers.

\section{Multi-Trajectory Diagnostic Benchmark}

\begin{figure}[h]
    \centering
    \includegraphics[width=1\linewidth]{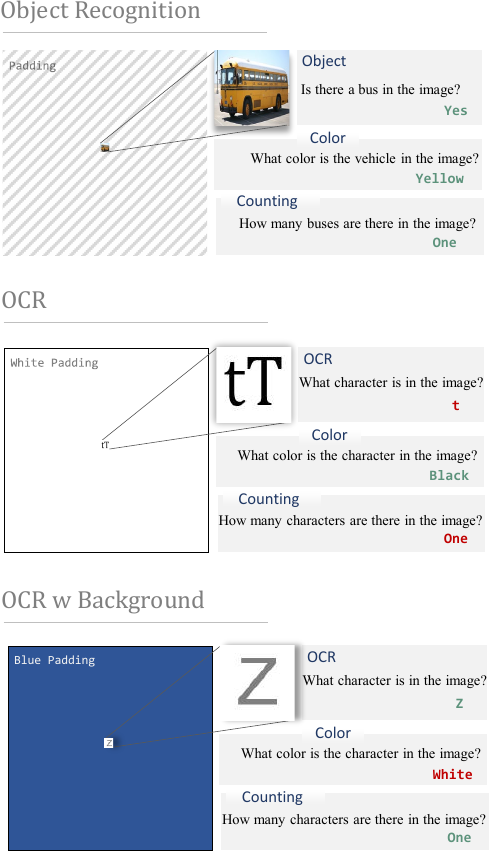}
    \caption{\textbf{Examples from the multi-trajectory benchmark.}
(Top) \textbf{Object recognition:} a bus image with associated object, color, and counting questions.
(Middle) \textbf{OCR:} a single character on white padding for text–color grounding.
(Bottom) \textbf{OCR with background:} same setup with a colored background, testing contextual bias in color reasoning.}
    \label{appfig:E_bench}
\end{figure}

To better illustrate the construction of our diagnostic benchmark introduced in~\cref{subsec:bench}, we provide detailed examples in \cref{appfig:E_bench}.
Each image is designed such that the target object or character occupies exactly one visual patch after resizing. For every image, we formulate three types of questions probing different semantic axes:
(1) Object recognition, verifying whether the model can correctly identify the target entity;
(2) Color identification, assessing grounding of visual color semantics; and
(3) Counting, testing numerical reasoning within the patch.

The benchmark comprises three subgroups:
\begin{enumerate}
    \item \textbf{Object}, object images with color and count variations;
    \item \textbf{OCR}, isolated characters placed on white padding, evaluating fine-grained recognition and color binding;
    \item \textbf{OCR with background}, characters placed on colored backgrounds, designed to measure contextual color bias and test whether the model grounds color to the object itself or to the dominant surrounding region.
\end{enumerate}

\section{Qualitative Analysis on Contextual Color Bias}
While our quantitative benchmark (see~\cref{subsec:bench}) reveals that MLLMs often confuse object and background colors, the underlying cause of this phenomenon remains unclear. In this section, we provide qualitative evidence demonstrating that such errors stem from \emph{contextual color bias}—a tendency for models to associate color semantics with dominant or surrounding visual regions rather than the object itself. Through controlled visual manipulations and \emph{EmbedLens} layer-wise tracing, we show that MLLMs infer color primarily from context statistics instead of grounded object cues.

\subsection{Bias Toward Dominant Patch Color}
As discussed in the “OCR w Background” examples of \cref{appfig:E_bench}, MLLMs frequently predict the dominant patch color rather than the actual object color when the two differ. This confirms that color reasoning within a patch is often driven by surface-level color statistics rather than grounded object understanding.

\subsection{Bias Toward Surrounding Color Context}
\begin{figure}[h]
    \centering
    \includegraphics[width=1\linewidth]{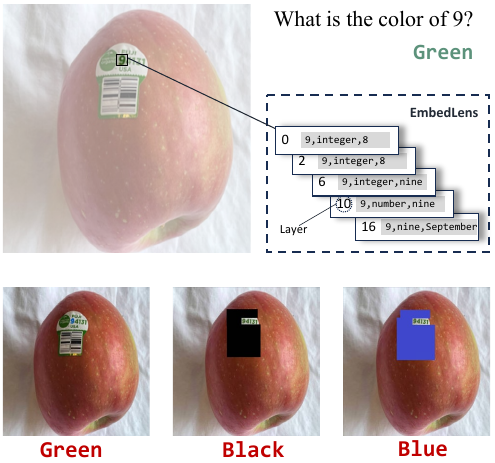}
    \caption{\textbf{Bias Toward Surrounding Color Context}}
    \label{appfig:F_contextual_bias_case_2}
\end{figure}

In the general OCR subset, most color errors correspond to the color of the surrounding padding instead of the target object. To verify this, we visualize the layer-wise decoding of \emph{EmbedLens} for the case shown in \cref{appfig:F_contextual_bias_case_2}.
Initially, the character “9” contains no green-related semantics at any layer, yet the model correctly answers “green.” When we recolor the digit itself (to blue), the model still predicts “green,” showing that it does not ground color to the digit. However, once we modify the color of the surrounding background, the prediction immediately follows the new context, confirming that \emph{color reasoning is largely inferred from nearby visual context rather than intrinsic object appearance}.

\section{Semantics Grounding for Vision Late Entry MLLMs}
To examine how delayed visual input affects semantic grounding, we analyze \emph{vision late-entry} MLLMs—models where visual tokens are introduced only at intermediate layers of the LLM backbone. One might expect that delaying visual injection allows the projector to take on stronger alignment responsibilities, yielding more semantically grounded visual tokens upon entry. However, our findings show that this is not the case.

\begin{figure}[h]
    \centering
    \includegraphics[width=1\linewidth]{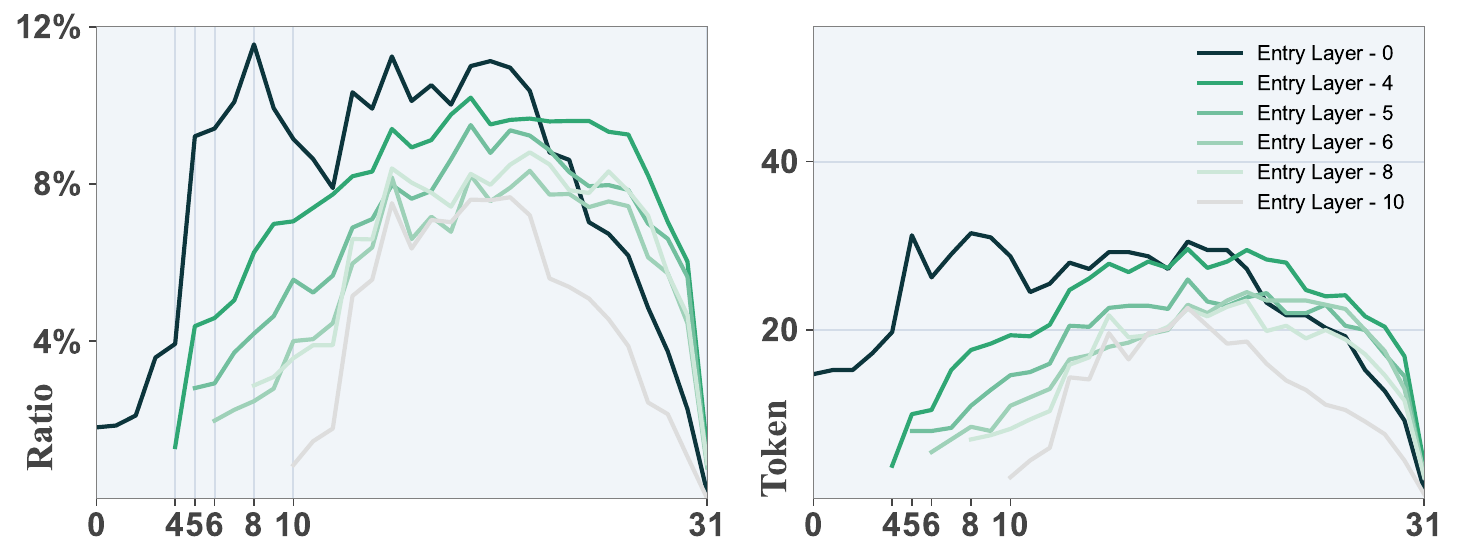}
    \caption{\textbf{Semantic grounding in vision late-entry models.}}
    \label{appfig:G}
\end{figure}

As shown in \cref{appfig:G}, the proportion of semantically grounded visual tokens at the entry point (left) remains nearly identical across entry layers (0–10), indicating that the projector itself does not inherently enhance semantic alignment. Instead, we observe that models with later visual entry continue to refine visual semantics over a broader range of layers, whereas dense models reach a stable alignment stage much earlier. This suggests that late-entry architectures shift the burden of cross-modal alignment deeper into the LLM, distributing semantic fusion across subsequent layers rather than concentrating it near the input.

\end{document}